\pdfoutput=1
\documentclass[graybox]{svmult}

\usepackage{mathptmx}       
\usepackage{helvet}         
\usepackage{courier}        
\usepackage{type1cm}        
\usepackage{makeidx}         
\usepackage{graphicx}        
\usepackage{multicol}        
\usepackage[bottom]{footmisc}

\makeindex             

\usepackage{listings}
\usepackage{float}
\usepackage{placeins}
\usepackage{amssymb}
\usepackage{amsmath}
\usepackage{lineno,hyperref}
\usepackage{subcaption}
\usepackage{multirow}
\usepackage{balance}
\usepackage{bm}

\usepackage{graphicx}

\usepackage{algorithm}
\usepackage[noend]{algpseudocode}

\makeatletter
\def\BState{\State\hskip-\ALG@thistlm}
\makeatother

\newcommand{\x}{\mathbf{x}}
\renewcommand{\u}{\mathbf{u}}
\newcommand{\y}{\mathbf{y}}
\newcommand{\W}{\mathbf{W}}
\newcommand{\Win}{\W_{in}}

\newcommand{\Wout}{\W_{out}}

\newcommand{\R}{\mathbb{R}}

\newcommand{\V}{\mathbf{V}}
\newcommand{\Vin}{\V_{in}}

\begin{document}

\title*{Hierarchical Temporal Representation in Linear Reservoir Computing}
\titlerunning{Hierarchical Temporal Representation in Linear RC}

\author{Claudio Gallicchio and Alessio Micheli and Luca Pedrelli}
\institute{Claudio Gallicchio \and Alessio Micheli \and Luca Pedrelli 
\at Department of Computer Science, University of Pisa, Largo B. Pontecorvo 3, Pisa, Italy\\
\email{{gallicch, micheli, luca.pedrelli}@di.unipi.it}
}

\maketitle
\abstract{Recently, studies on deep Reservoir Computing (RC) highlighted the role of 
layering in deep recurrent neural networks (RNNs). In this paper, the use of linear recurrent units allows
us to bring more evidence on the intrinsic hierarchical temporal representation in deep RNNs 
through frequency analysis applied to the state signals.
The potentiality of our approach is assessed on the class of Multiple Superimposed Oscillator 
tasks. Furthermore, our investigation provides useful insights to open a discussion on 
the main aspects that characterize the deep learning framework in the temporal domain.
}

\keywords{Reservoir Computing, Deep Learning, Deep Echo State Network, Multiple Time-Scales Processing.}

\section{Introduction}

In the last years, the extension of deep neural network architectures
towards recurrent processing of temporal data has opened the way 
to novel approaches to effectively learn 
hierarchical representations of time-series featured by multiple time-scales dynamics
\cite{Schmidhuber2015, Pascanu2014, ElHihi1995, Hermans2013, Angelov2016}.
Recently, within the umbrella of randomized neural network approaches \cite{Gallicchio2017Randomized}, 
Reservoir Computing (RC) \cite{verstraeten2007experimental, lukovsevivcius2009reservoir}
has proved to be a useful  tool for analyzing the intrinsic properties of stacked architectures in recurrent neural networks (RNNs), allowing at the same time to exploit the extreme efficiency of RC training algorithms 
in the design of novel deep RNN models. 
Stemming from the Echo State Network (ESN) approach \cite{Jaeger2004}
the study of the dynamics of multi-layered recurrent reservoir architectures has been 
introduced with the deepESN model in \cite{gallicchio2017deep,Gallicchio2016Deep}.
In particular, the outcomes of the experimental analysis in \cite{gallicchio2017deep,Gallicchio2016Deep} as well as theoretical results in the field of dynamical systems \cite{gallicchio2017echo,Gallicchio2017Lyapunov}, highlighted the role of layering in the inherent development of progressively more abstract temporal representations
in the higher layers of deep recurrent models.

In this paper, we take a step forward in the study of the structure of the temporal features 
naturally emerging in layered RNNs. To this aim, we resort to classical tools in the area of signal processing to analyze the differentiation among the  state representations developed by the different levels of a deepESN in a task involving signals in a controlled scenario. In particular, we simplify the deepESN design by implementing recurrent units with linear activation function, i.e. we adopt linear deepESN (L-deepESN). 
In the analysis of the frequency spectrum of network's states, this approach brings the major advantage of 
avoiding the effects of harmonic distortion due to non-linear activation functions. To provide a 
quantitative support to our analysis, we  experimentally assess the L-deepESN model on a variety of progressively more involving versions of the Multiple Superimposed Oscillator (MSO) task \cite{wierstra2005modeling, xue2007decoupled}.
Note that the class of MSO tasks is of particular interest for the aims of this paper,
especially in light of previous literature results that pointed out the relevant need for multiple time-scales processing ability \cite{Jaeger2007, schmidhuber2007training, xue2007decoupled} as well as the potentiality of linear models in achieving excellent predictive results in base settings of the problem \cite{vcervnansky2008predictive}.
Another example of application of linear RNNs is in \cite{pasa2014pre}.

As a further contribution, our investigation would offer interesting insights on the  nature of compositionality in deep learning architectures. Typically, deep neural networks consist in a hierarchy of many non-linear hidden layers 
that enable a distributed information representation (through learning) where higher layers specialize to  progressively more abstract concepts. Removing the  characteristic of non-linearity, and focusing on the ability to develop a hierarchical diversification of temporal features (prior to learning), our analysis sheds new light into the true essence of layering in 
deep RNN  even with \emph{linear} recurrent units.

The rest of this paper is organized as follows. In Section~\ref{sec:linearDeepESN} we introduce the L-deepESN model. 
In Section~\ref{sec:exp} we analyze the hierarchical nature of temporal representations in L-deepESN, presenting the experimental results on the MSO tasks and the outcomes of the signal processing analysis of the developed system dynamics.
Finally, in Section~\ref{sec:Conclusions} we draw the conclusions.

\section{Linear Deep Echo State Networks}
\label{sec:linearDeepESN}
\begin{figure}[htb]
\sidecaption
\includegraphics[width = \textwidth]{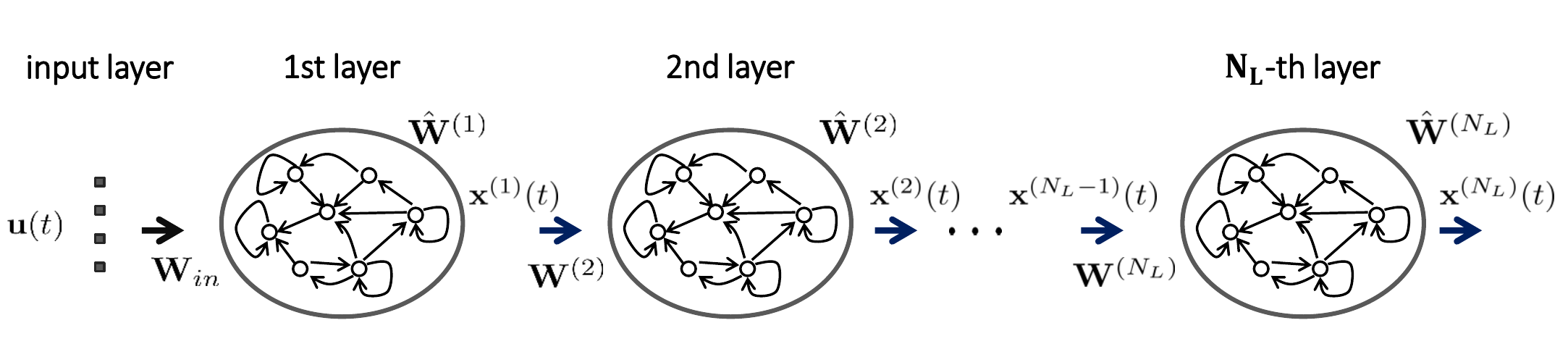}
\caption{Layered reservoir architecture in a deepESN.}.
\label{fig.deepESN}
\end{figure}

A deepESN architecture \cite{gallicchio2017deep} is composed by a stack of $N_L$ recurrent reservoir layers, where 
at each time step $t$ the first layer receives the external input $\u(t) \in \R^{N_U}$, 
while successive layers are fed by the output of the previous layer in the hierarchy.
We denote the state  of layer $i$ at time $t$ by $\x^{(i)}(t) \in \R^{N_R}$,
where we assume the same state dimension $N_R$
for every layer for the sake of simplicity.
A schematic representation of the reservoir architecture in a deepESN
is provided in Figure~\ref{fig.deepESN}.

By referring to the case of leaky integrator reservoir units \cite{Jaeger2007}, and omitting the bias terms for the ease of notation, the state transition function of the first layer
is given by the following equation:
\begin{equation}
\label{eq.layer1}
\x^{(1)}(t) = (1-a^{(1)}) \x^{(1)}(t-1) + a^{(1)} \mathbf{f}(\Win \u(t) + \hat{\W}^{(1)} \x^{(1)}(t-1)),
\end{equation}
whereas the state transition of layer $i>1$ is ruled by the equation:
\begin{equation}
\label{eq.layeri}
\x^{(i)}(t) = (1-a^{(i)}) \x^{(i)}(t-1) + a^{(i)} \mathbf{f}(\W^{(i)} \x^{(i-1)}(t) + \hat{\W}^{(i)} \x^{(i)}(t-1)),
\end{equation}
where $a^{(i)} \in [0,1]$ is the leaking rate parameter at layer $i$, 
$\Win \in \R^{N_R \times N_U}$ is the input weight matrix,
$\W^{(i)} \in \R^{N_R \times N_R}$ is the weight matrix of the inter-layer connections from layer $i-1$ to layer $i$,
$\hat{\W}^{(i)} \in \R^{N_R \times N_R}$ is the matrix of recurrent weights of layer $i$, 
and $\mathbf{f}$ denotes the element-wise application of the activation function of the recurrent units.
A null initial state is considered for the reservoirs in all the layers, i.e. $\x^{(i)}(0) = \mathbf{0}$ for all $i = 1, \ldots, N_L$.

The case of L-deepESN  is obtained from equations~\ref{eq.layer1} and \ref{eq.layeri}
when a linear activation function is used for each recurrent unit, i.e. $\mathbf{f} = \mathbf{id}$.
As in standard RC, all the reservoirs parameters, i.e.  all the weight matrices in equations
\ref{eq.layer1} and \ref{eq.layeri},
are left untrained after initialization subject to stability constraints.
According to the necessary condition for the Echo State Property of deep RC networks 
\cite{gallicchio2017echo}, stability can be accomplished by constraining the maximum among the 
spectral radii of  matrices $\left((1-a^{(i)}) \mathbf{I} + a^{(i)}\hat{\W}^{(i)}\right)$, individually denoted by $\rho^{(i)}$, to be not above unity.
Thereby, a simple initialization condition for L-deepESNs consists in randomly selecting the weight values in 
matrices $\W_{in}$ and $\{\W^{(i)}\}_{i = 2}^{N_L}$  from a uniform distribution in $[-scale_{in}, scale_{in}]$,
whereas the weights in recurrent matrices $\{\hat{\W}^{(i)}\}_{i=1}^{N_L}$ are initialized in a similar way and are then re-scaled to meet the condition on $\max{\rho^{(i)}}$.

In this context it also interesting to observe that the use of linearities 
allows us to express the evolution of the whole system  
by means of an algebraic expression
that describes the dynamics 
of an equivalent single-layer recurrent system with the same total number
of recurrent units.
Specifically, denoting by $\x(t) = (\x^{(1)}(t), \x^{(2)}(t), \ldots, \x^{(N_L)}(t)) \in \R^{N_L N_R}$ the global state of the network,
the dependence of $\x(t)$ from $\x(t-1)$ can be expressed as $\x(t) = \V \x(t-1) + \V_{in} \u(t)$,
where both $\V\in\R^{N_L N_R \times N_L N_R}$ and $\V_{in}\in \R^{N_L N_R \times N_U}$ can be viewed as block matrices, with block elements denoted respectively by $\V_{i,j} \in \R^{N_R \times N_R}$ and 
$\V_{in,i} \in \R^{N_R \times N_U}$, i.e.:
\begin{equation}
\label{eq.whole}
\x(t) = 
\left[
\begin{array}{lll}
\V_{1,1} & \ldots & \V_{1,N_L} \\
\vdots & \ddots & \vdots \\
\V_{N_L,1} & \ldots & \V_{N_L,N_L} \\
\end{array}
\right]
\x(t-1) + 
\left[
\begin{array}{l}
\V_{in,1} \\
\vdots \\
\V_{in,N_L}
\end{array}
\right]
\u(t).
\end{equation}
Noticeably, the layered organization imposes a lower triangular block matrix structure to $\V$
such that in the linear case its blocks can be computed as:
\begin{equation}
\label{eq.W}
\V_{i,j} = 
\left\lbrace
\begin{array}{ll}
\mathbf{0} & \text{if } i < j \\ 
(1-a^{(i)})\mathbf{I}+a^{(i)}\hat{\W}^{(i)} & \text{if } i = j \\
(\prod_{k = j+1}^{i} a^{(k)}{\W}^{(k)}) \big((1-a^{(j)})\mathbf{I}+a^{(j)}\hat{\W}^{(j)}\big) \quad\quad & \text{if } i > j. \\
\end{array}
\right.
\end{equation}
Moreover, as concerns the input matrix, we have:
\begin{equation}
\label{eq.Win}
\V_{in,i} = 
\left\lbrace
\begin{array}{ll}
a^{(1)} \W_{in} & \text{if } i = 1 \\
(\prod_{k=2}^i a^{(k)} \W^{(k)}) a^{(1)} \W_{in} \quad \quad & \text{if } i > 1. \\
\end{array}
\right.
\end{equation}

The mathematical description provided here for the L-deepESN case is particularly helpful in order to highlight the characterization resulting from the layered composition of recurrent units. Indeed, from an architectural perspective, 
a deep RNN can be seen as obtained by imposing a set of constraints to the architecture of a single-layer fully connected RNN with the same total number of recurrent units. Specifically, a deep RNN can be obtained from the architecture of a fully connected (shallow) RNN by removing the recurrent connections corresponding in the deep version to the connections
from higher layers to lower layers, as well as the input connections to the levels higher than 1.
In this respect, the use of linear activation functions has the effect of enhancing the emergence of such 
constrained characterization and making it visible through the peculiar algebraic organization of the state update 
as described by equations~\ref{eq.whole}, \ref{eq.W} and \ref{eq.Win}. 
Indeed, the constrained structure given by the layering factor is reflected in the (lower triangular block) structure of the 
matrix $\V$ that rules  the recurrence of the whole network dynamics in equation~\ref{eq.whole}.
In particular, the last line of equation~\ref{eq.W} highlights the progressive filtering effect on the state information propagated towards the higher levels in the network, modulated by the leaking rates and through the magnitude of the inter-layer weights values. Similarly, the last line of equation~\ref{eq.Win} shows the analogous progressive filtering effect operated on the external input information for increasing level's depth.

Thereby, although from the system dynamics viewpoint 
it is possible to find a shallow recurrent network that is equivalent
to an L-deepESN, the resulting form of the  matrices that rules the state evolution, i.e. $\V$ and $\Vin$,
has a distinct characterization that is due to the layered construction.
Moreover, note that the probability of obtaining such matrices $\V$ and $\Vin$ 
by means of standard random reservoir initialization is negligible.
Noteworthy, the aforementioned architectural constraints imposed by the hierarchical construction are reflected also in the ordered structure of the temporal features represented in higher levels of the recurrent architecture, 
as investigated for linear reservoirs in Section~\ref{sec:exp},
and as observed, under a different perspective and 
using different mathematical tools, 
in the non-linear case in \cite{gallicchio2017deep}.

As regards network training, as in standard RC, the only learned parameters of the L-deepESN are those pertaining to the 
readout layer. This is used for output computation by means of a linear combination 
of the reservoir units activations in all the levels, 
allowing the linear learner to weight differently the
contributions of the multiple dynamics developed in the network state.
In formulas, at each time step $t$ the output $\y(t) \in \R^{N_Y}$ is computed as $\y(t) = \Wout \x(t)$, where
$\Wout \in \R^{N_Y \times N_L N_R}$ is the output weight matrix whose values are learned from a training set. 
Typically, as in the standard RC framework, the values in $\Wout$ are found in closed form by using direct methods such as pseudo-inversion or ridge regression \cite{lukovsevivcius2009reservoir}.

\section{Experimental Assessment}
\label{sec:exp}

In this section we present the results of the experimental assessment of L-deepESN on the class of MSO tasks.

\begin{figure}[h!] 
    \centering
    \includegraphics[width=0.7\linewidth]{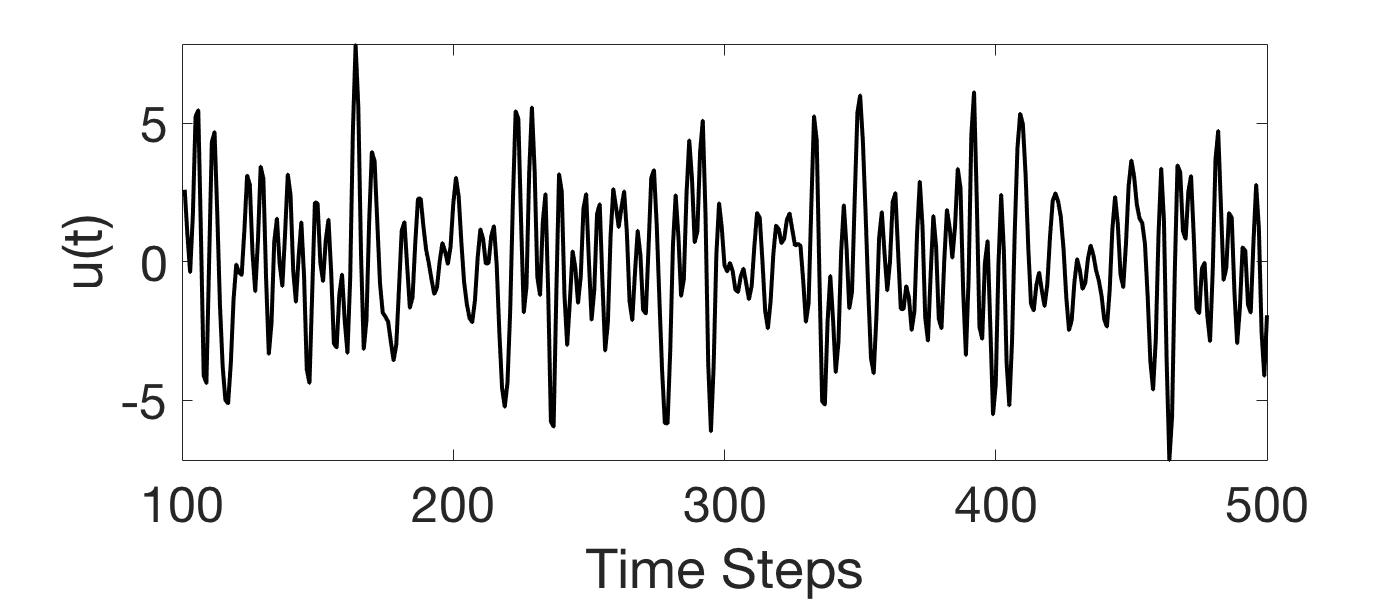} 
 \caption{A 400 time step long excerpt of the input sequence for the MSO12 task.} 
 \label{fig:mso12} 
\end{figure}

An MSO task consists in a next-step prediction on a 1-dimensional time-series, 
i.e. for each time step $t$ the target output is given by $y_{target}(t) = u(t+1)$, where
$N_U = N_Y = 1$.
The considered time-series is given by a sum of sinusoidal functions:
\begin{equation}
\label{eq:input}
u(t)= \sum_{i=1}^n sin(\varphi_i t)
\end{equation}
where $n$ denotes the number of sinusoidal functions, $\varphi_i$ determines the frequency of the $i$-th sinusoidal function and $t$ is the index of the time step. 
In the following, we use the notation MSO$n$ to 
specify  the number $n$ of sinusoidal functions
that are accounted in the task definition.
The $\varphi_i$ coefficients in equation~\ref{eq:input}
are set as in \cite{otte2016optimizing, koryakin2012balanced}, i.e.
$\varphi_1 = 0.2, \varphi_2 = 0.331, \varphi_3 = 0.42, \varphi_4 = 0.51, \varphi_5 = 0.63, \varphi_6 = 0.74, \varphi_7 = 0.85, \varphi_8 = 0.97, \varphi_9 = 1.08, \varphi_{10} = 1.19, \varphi_{11} = 1.27, \varphi_{12} = 1.32$. 
In particular, in our experiments we focus on versions of the MSO task
with a number of sine waves $n$ ranging from 5 to 12.
This allows us to exercise the ability of the RC models to 
develop a hierarchy of temporal representations in challenging cases where the input signal is enriched by the
presence of many different time-scales dynamics. Besides, note that summing an increasing number of sine waves with frequencies that are not integer multiples of each other makes the prediction task  harder due to the increasing signal period.
An example of the input signal for the MSO12 task is given in Figure~\ref{fig:mso12}.
For all the considered settings of the MSO task, the first 400 steps are used for training (with a washout of length 100),
time steps from 401 to 700 are used for validation and the remaining steps from 701 to 1000 are used for test.

In our experiments, we used L-deepESN with $N_L$ levels, each consisting in 
a fully connected reservoir with $N_R$ units. We assumed that $\W_{in}$ and $\{\W^{(i)}\}_{i = 2}^{N_L}$ are initialized with the same scaling parameter  $scale_{in}$, 
and we used the same value of the spectral radius and of the leaking rate in every level,
i.e. $\rho^{(i)} = \rho$  and $a^{(i)} = a$ for every $i$. For readout training we used ridge regression.
Table~\ref{tab:hyper_parameters} reports the range of values considered for every hyper-parameter considered in our experiments.
\begin{table}[H]
\begin{center}
\begin{tabular}{|l|l|}
\hline
\emph{Hyper-parameter} & \emph{Values considered for model selection} \\
\hline
\hline
number of levels $N_{L}$ & 10 \\
\hline
reservoir size $N_{R}$ & 100 \\
\hline
input scaling $scale_{in}$ & 0.01, 0.1, 1 \\
\hline
leaking rate $a$ & 0.1, 0.3, 0.5, 0.7, 0.9, 1.0 \\
\hline
spectral radius $\rho$ & 0.1, 0.3, 0.5, 0.7, 0.9, 1.0 \\
\hline
ridge regression regularization $\lambda_r$ & $10^{-11}, 10^{-10}, ..., 10^0$ \\
\hline
\end{tabular}
\end{center}
\caption{Hyper-parameters values considered for model selection on the MSO tasks.}
\label{tab:hyper_parameters}
\end{table}

In order to evaluate the predictive performance on the MSO tasks, 
we used the normalized root mean square error (NRMSE), calculated as follows:
\begin{equation}
NRMSE = \sqrt{(\sum_{t=1}^{T} (y_{target}(t)-y(t))^2) / (T\sigma_{y_{target}(t)}^2}),
\end{equation}
where $T$ denotes the sequence length, $y_{target}(t)$ and $y(t)$ are the target and the network's output at time $t$, and $\sigma_{y_{target}(t)}^2$ is the variance of $y_{target}$. For each reservoir hyper-parametrization, we independently generated 10 reservoir guesses, the predictive performance in the different cases has been averaged over such guesses and then the 
model's hyper-parameterization has been selected on the validation set.

In the following Sections \ref{sec:results} and \ref{sec:frequency} we respectively  evaluate our approach from 
a quantitative point of view, comparing the predictive performance of L-deepESN with 
related literature models, and from a  qualitative perspective, by analyzing the frequencies of the state activations developed in the different reservoir levels.

\subsection{Predictive Performance}
\label{sec:results}
In this section 	 
on recent (more complex and richer) variants of the MSO task, 
with a number of sine waves $n$ varying from 5 to 12.
Table \ref{tab:models_comparison} provides a comparison among the 
NRMSE achieved on the test set by L-deepESN, 
neuro-evolution \cite{otte2016optimizing}, balanced ESN \cite{koryakin2012balanced}, 
ESN with infinite impulse response units (IIR ESN)  \cite{holzmann2010echo} and Evolino \cite{schmidhuber2007training} on the considered MSO tasks.
Furthermore, in the same table, we report the performance achieved by 
linear ESN built with a single fully connected 
reservoir (L-ESN), 
considering the same range of hyper-parameters and total number of recurrent units
as in the L-deepESN case.

\begin{table}[H]
\centering
\begin{tabular}{|l|c|c|c|c|c|c|c|}
\hline
\emph{Task} & L-deepESN & L-ESN & n.-evolution \cite{otte2016optimizing} & balanced ESN \cite{koryakin2012balanced} & IIR ESN \cite{holzmann2010echo}  & Evolino \cite{schmidhuber2007training} \\
\hline
\hline
MSO5 & $6.75 \cdot 10^{-13}$  & $7.14 \cdot 10^{-10}$  & $4.16 \cdot 10^{-10}$  & $1.06 \cdot 10^{-6}$ & $8\cdot 10^{-5}$ & $1.66 \cdot 10^{-1}$   \\
\hline
MSO6 & $1.68 \cdot 10^{-12}$  & $5.40\cdot 10^{-9}$ & $9.12\cdot 10^{-9}$ & $8.43\cdot 10^{-5}$ & - & -    \\
\hline
MSO7 & $5.90\cdot 10^{-12}$  & $5.60\cdot 10^{-8}$ & $2.39\cdot 10^{-8}$ & $1.01\cdot 10^{-4}$ &- &  - \\
\hline
MSO8 & $1.07\cdot 10^{-11}$  & $2.08\cdot 10^{-7}$ & $6.14\cdot 10^{-8}$ & $2.73\cdot 10^{-4}$ &-  & -  \\
\hline
MSO9 & $5.34\cdot 10^{-11}$  & $4.00\cdot 10^{-7}$ & $1.11\cdot 10^{-7}$ &-  & - & -  \\
\hline
MSO10 & $8.22\cdot 10^{-11}$  & $8.21\cdot 10^{-7}$ & $1.12\cdot 10^{-7}$ &- & -&  -   \\
\hline
MSO11 & $4.45\cdot 10^{-10}$  & $1.55\cdot 10^{-6}$ & $1.22\cdot 10^{-7}$ &- &- & -    \\
\hline
MSO12 & $5.40\cdot 10^{-10}$  & $1.70\cdot 10^{-6}$ & $1.73\cdot 10^{-7}$ &- &- &- \\
\hline
\end{tabular}
\caption{Test NRMSE obtained 
by L-deepESN, L-ESN, neuro-evolution (n.-evolution), balanced ESN, IIR ESN and Evolino on 
the MSO5-12 tasks.}
\label{tab:models_comparison}
\end{table}

Noteworthy, the proposed L-deepESN model outperformed the best literature results of about 3 or 4 orders of 
magnitude on all the MSO settings. Furthermore, test errors obtained by L-ESN are always 
within one order of magnitude of difference with respect 
to the best state-of-the-art results. 
These aspects confirms the effectiveness of the linear activation function on this task,
as also testified by our preliminary results that showed 
poorer performance for RC networks with $tanh$ units, unless forcing the operation of the activation function
in the linear region.
Moreover, L-deepESN always performed  better then L-ESN. 
On the basis of the known characterization of the MSO task,
our results confirm the quality of the hierarchical structure of recurrent reservoirs in representing multiple time-scales dynamics with respect to its shallow counterpart.

For the sake of completeness, we performed a further comparison considering L-deepESNs with the same number of total recurrent units  used by the other ESN models taken into account
from literature. 
In particular, balanced ESN used a maximum number of 250 units for model selection on the MSO5, MSO6, MSO7 and MSO8 tasks,
while IIR ESN implemented 100 units on the MSO5 task (see results in Table~\ref{tab:models_comparison}). 
L-deepESN with $N_L = 10$ and $N_R = 25$  (i.e. a total of 250 recurrent units) performed better than balanced ESN, obtaining a test NRMSE of $1.20 \cdot 10^{-11}$, $8.73 \cdot 10^{-11}$, $2.42 \cdot 10^{-10}$ and $9.06 \cdot 10^{-10}$, on the MSO5, MSO6, MSO7 and MSO8 tasks, respectively. Moreover, even L-deepESN with $N_L = 10$ and $N_R = 10$  (i.e. a total of 100 recurrent units) obtained a better performance than IIR ESN, achieving a test error of $7.41 \cdot 10^{-11}$ on the MSO5 task.

\subsection{Hierarchical Temporal Representation Analysis}
\label{sec:frequency}
In this section we investigate the temporal representation developed by the reservoirs levels in an L-deepESN, 
using as input signal the sequence considered for the MSO12 task, featured by rich dynamics with known multiple time-scales characterization (see equation~\ref{eq:input}).
We used the same reservoir hyper-parameterization selected for the predictive experiments on the MSO12 task in Section~\ref{sec:results}, 
namely $N_R = 100$, $N_L = 10$, $scale_{in} = 1$, $a = 0.9$ and $\rho = 0.7$,
averaging the results over 100 reservoir guesses.
In our analysis, we first computed the states obtained by running the L-deepESN on the input sequence.
Then, we performed the Fast Fourier Transform (FFT) \cite{frigo1998fftw} algorithm on 
the states of all the recurrent units over the time, normalizing the obtained values in order to enable a qualitative comparison. Finally, we averaged the FFT values on a layer-by-layer basis. 

\begin{figure}[htp] 
    \centering
    \includegraphics[width=0.45\linewidth]{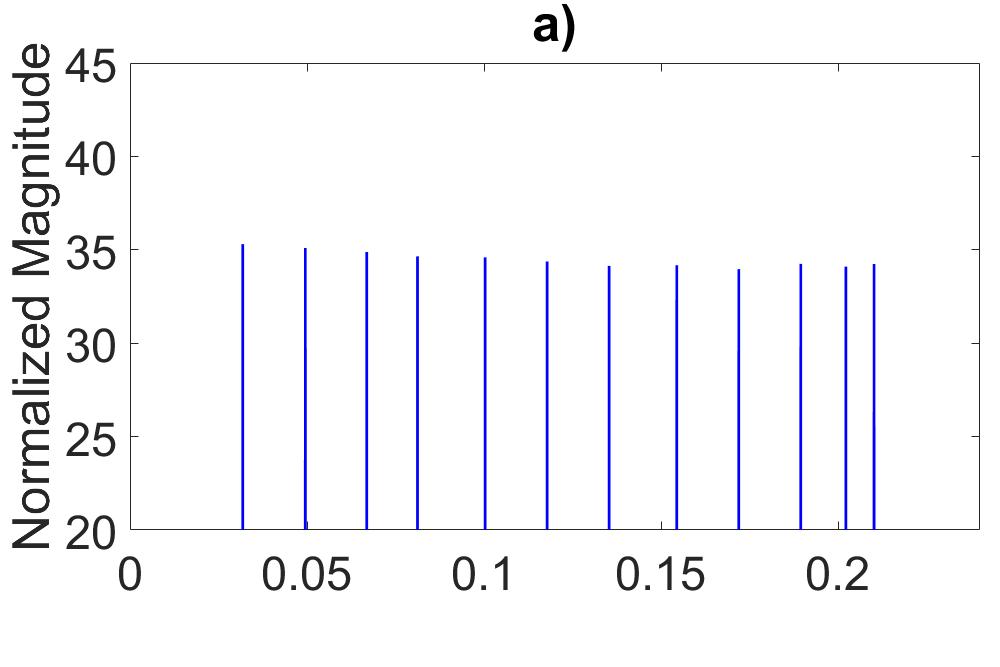} 
    \includegraphics[width=0.45\linewidth]{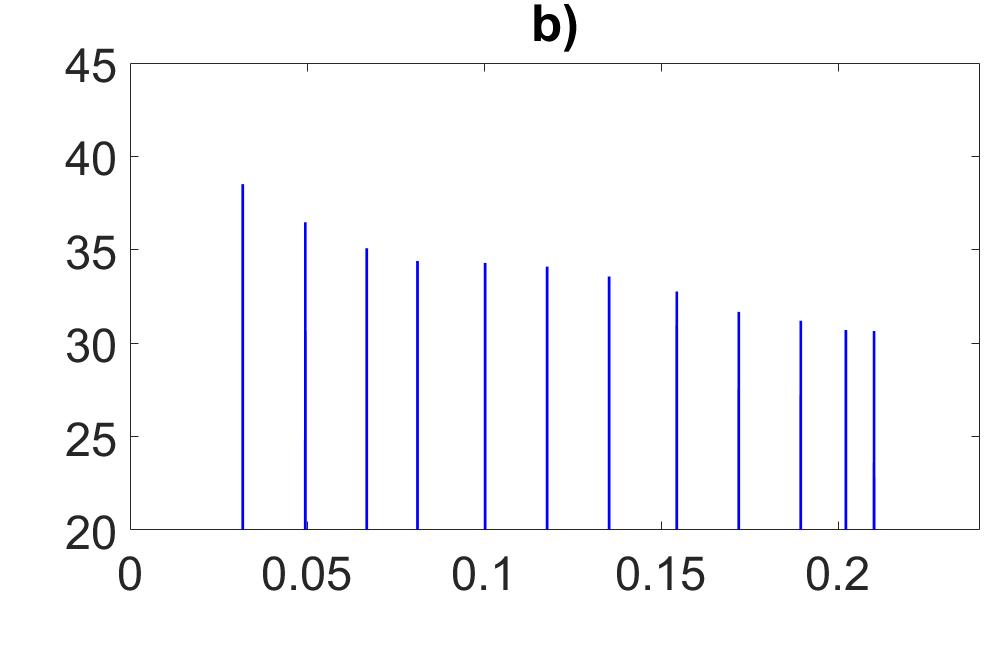} \\    
    \includegraphics[width=0.45\linewidth]{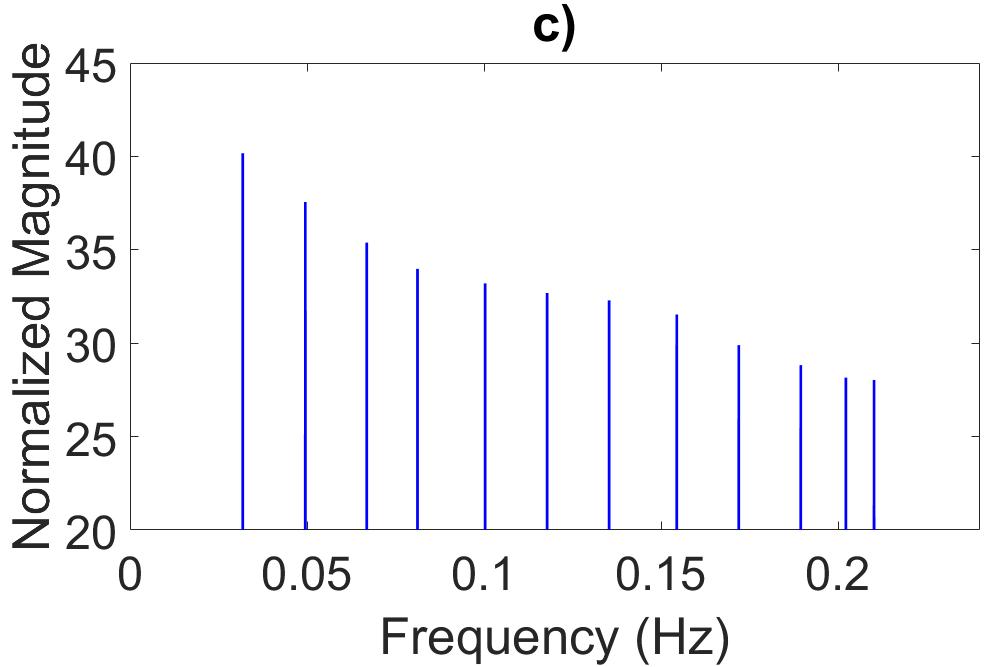} 
    \includegraphics[width=0.45\linewidth]{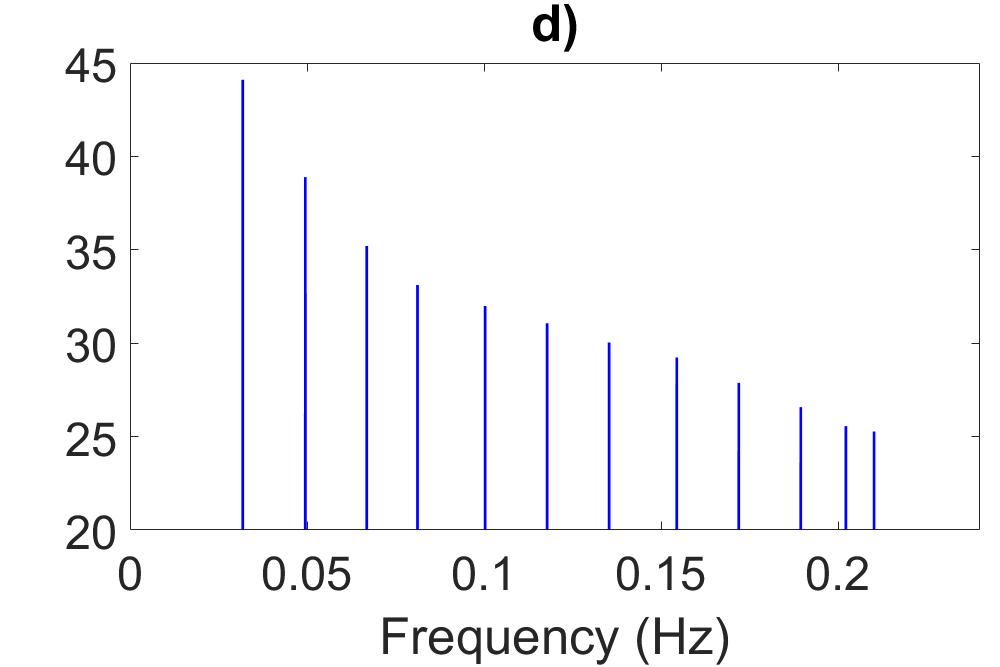} 
     \caption{FFT components of reservoir states in progressively higher levels of L-deepESN,
              \textbf{a}): level 1, \textbf{b}): level 4, \textbf{c}): level 7,
              \textbf{d}): level 10.}
 \label{fig:frequency} 
\end{figure}

The FFT values obtained for progressively higher levels of  L-deepESN are shown in 
Figures \ref{fig:frequency}\textbf{a}), \ref{fig:frequency}\textbf{b}), \ref{fig:frequency}\textbf{c}) and \ref{fig:frequency}\textbf{d}), which respectively focus on levels $1$, $4$, $7$ and $10$.
These figures represent the state signal in the frequency domain,
where it is possible to see  12 spikes corresponding to the 12 sine waves components of the input.
Looking at the magnitude of the FFT components, i.e. at the height of the spikes in plots,
we can have an indication of how the signals are elaborated by  the individual recurrent levels.
We can see that the state of the reservoir at level 1 shows FFT components all with approximately the same magnitude.
The FFT components of reservoir states at levels 4, 7 and 10, instead, show diversified magnitudes. 
Specifically, we can see that in  higher levels of the network higher frequency components are progressively filtered,
and lower frequency components tend to have relative higher magnitudes. This confirms the insights on the progressive filtering effect discussed  in Section~\ref{sec:linearDeepESN} in terms of mathematical characterization of the system.

Results in Figure~\ref{fig:frequency} show that the hierarchical construction of recurrent models 
leads, even in the linear case, to a representation of the temporal signal that is  sparsely distributed across the 
network, where different levels tend to focus on a different range of frequencies.
Moreover, the higher is the level, the stronger is the focus on lower frequencies, hence the
state signals emerging in deeper levels are naturally featured  by coarser time-scales and slower dynamics.
Thereby, the layered organization of the recurrent units determines a temporal representation 
that has an intrinsic hierarchical structure. According to  this, the multiple time-scales in the network dynamics
are ordered depending to the depth of reservoirs' levels. Such inherent characterization of the  hierarchical distributed temporal representation can be  exploited  when training the readout, as testified by the excellent predictive performance of L-deepESN on the MSO tasks reported in Section~\ref{sec:results}.

\section{Conclusions}
\label{sec:Conclusions}

In this paper, we have studied the inherent properties of hierarchical linear RNNs by
analyzing the frequency of the states signals emerging in the different levels of the
recurrent architecture. The FFT analysis revealed that the stacked composition of reservoirs in a L-deepESN
tends to develop a structured representation of the temporal information.
Exploiting an incremental filtering effect, states in higher levels of the hierarchy 
are biased towards slower components of the frequency spectrum, resulting in progressively slower temporal dynamics.
In this sense, the emerging structure of L-deepESN states can be seen as an echo of the multiple time-scales present in the input signal, distributed across the layers of the network.
The hierarchical representation of temporal features in L-deepESN 
has been exploited to address recent challenging versions of the MSO task.
Experimental results showed that the proposed approach dramatically outperforms the state-of-the-art on the MSO tasks,
emphasizing the relevance of the hierarchical temporal representation  and also confirming the effectiveness of linear signal processing on the MSO problem.

Overall, we showed a concrete evidence that layering is an aspect of the network construction that is intrinsically able to provide a distributed and hierarchical feature representation of temporal data.
Our analysis pointed out that this is possible even without (or prior to) learning of the recurrent connections,
and releasing the requirement for non-linearity of the activation functions.
We hope that the considerations delineated in this paper could contribute to open an intriguing research question regarding the merit of shifting the focus, from the concepts of learning and non-linearities, to the concepts of hierarchical 
organization and distribution of representation to define the salient aspects of the deep learning framework for recurrent architectures.

\bibliographystyle{spmpsci}
\bibliography{references}

\end{document}